\documentclass[a4paper, 10pt, conference]{ieeeconf}      % Use this line for a4
\usepackage{FG2024}

\FGfinalcopy % *** Uncomment this line for the final submission

\IEEEoverridecommandlockouts                              % This command is only
\usepackage{booktabs}
\usepackage{hyperref}
\usepackage{cite}
\usepackage{graphicx}
\usepackage{multirow}
\usepackage{subcaption}

\def\FGPaperID{****} % *** Enter the FG2024 Paper ID here

\title{\LARGE \bf
SDFR: Synthetic Data for Face Recognition Competition
}

\author{\parbox{16cm}{\centering
    {\large Hatef Otroshi Shahreza$^{1,2,\dagger}$, 
    	Christophe Ecabert$^{1,\dagger,*}$, 
    	Anjith George$^{1,\dagger,*}$,\\
    	Alexander Unnervik$^{1,2,\dagger,*}$, Sébastien Marcel$^{1,3,\dagger}$,
    	Nicolò Di Domenico$^{4,\ddagger}$, \\
    	Guido Borghi$^{4,\ddagger}$, 
    	Davide Maltoni$^{4,\ddagger}$, 
    	Fadi Boutros$^{5,\ddagger}$, 
    	Julia Vogel$^{5,\ddagger}$, \\  
    	Naser Damer$^{5,6,\ddagger}$,   
    	Ángela Sánchez-Pérez$^{7,8,\ddagger}$,
    	Enrique Mas-Candela$^{7,8,\ddagger}$, \\ 
    	Jorge Calvo-Zaragoza$^{7,\ddagger}$, 
    	Bernardo Biesseck$^{9,10,\ddagger}$, 
    	Pedro Vidal$^{9,\ddagger}$, 
    	Roger Granada$^{11,\ddagger}$, \\  
    	David Menotti$^{9,\ddagger}$, 
    	Ivan DeAndres-Tame$^{12,\ddagger}$, 
    	Simone Maurizio La Cava$^{13,\ddagger}$, 
    	Sara Concas$^{13,\ddagger}$, 
    	Pietro Melzi$^{12,\ddagger}$, 
    	Ruben Tolosana$^{12,\ddagger}$, 
    	Ruben Vera-Rodriguez$^{12,\ddagger}$, 
    	Gianpaolo Perelli$^{13,\ddagger}$, \\  
    	Giulia Orrù$^{13,\ddagger}$, 
    	Gian Luca  Marcialis$^{13,\ddagger}$, 
    	Julian Fierrez$^{12,\ddagger}$
    }\\
    {\normalsize
    	\vspace{5pt}
    $^{1}$Idiap Research Institute, Switzerland  
    $^{2}$École Polytechnique Fédérale de Lausanne (EPFL),  Switzerland 
    $^{3}$Université de Lausanne (UNIL), Switzerland 
    $^{4}$University of Bologna, Italy \\
    $^{5}$Fraunhofer Institute for Computer Graphics Research (IGD), Germany  
    $^{6}$TU Darmstadt, Germany  \\
    $^{7}$Universidad de Alicante, Spain, 
    $^{8}$Facephi Biometria SA, R\&D Centre, Spain,    \\
    $^{9}$Federal University of Paraná (UFPR), Curitiba, PR, Brazil  \\
    $^{10}$Federal Institute of Mato Grosso (IFMT), Pontes e Lacerda, Brazil 
     $^{11}$unico - idTech, Brazil \\
     $^{12}$Universidad Autonoma de Madrid (UAM), Spain  
     $^{13}$Università degli Studi di Cagliari (UNICA), Italy 
    \\\vspace{5pt}
    $^{\dagger}$Competition organizer \quad $^{\ddagger}$Competition participant \\   $^{*}$Equal Contribution\\ \vspace{4pt}
}}
\thanks{ Competition Website: \href{http://www.idiap.ch/challenge/sdfr/}{www.idiap.ch/challenge/sdfr}}
}

\begin{document}

\ifFGfinal
\thispagestyle{empty}
\pagestyle{empty}
\else
\author{Anonymous FG2024 submission\\ Paper ID \FGPaperID \\}
\pagestyle{plain}
\fi

\maketitle

%%%%%%%%%%%%%%%%%%%%%%%%%%%%%%%%%%%%%%%%%%%%%%%%%%%%%%%%%%%%%%%%%%%%%%%%%%%%%%%%
\begin{abstract}
Large-scale face recognition datasets are collected by crawling the Internet and without individuals' consent, raising legal, ethical, and privacy concerns. With the recent advances in generative models, recently several works proposed generating synthetic face recognition datasets to mitigate concerns in web-crawled face recognition datasets. This paper presents the summary of the Synthetic Data for Face Recognition (SDFR) Competition held in conjunction with the 18th IEEE International Conference on Automatic Face and Gesture Recognition (FG 2024) and established to investigate the use of synthetic data for training face recognition models. The SDFR competition was split into two tasks, allowing participants to train face recognition systems using new synthetic datasets and/or existing ones. In the first task, the face recognition backbone was fixed and the dataset size was limited, while the second task provided almost complete freedom on the model backbone, the dataset, and the training pipeline. The submitted models were trained on existing and also new synthetic datasets and used clever methods to improve training with synthetic data. The submissions were evaluated and ranked on a diverse set of seven benchmarking datasets. The paper gives an overview of the submitted face recognition models and reports achieved performance compared to baseline models trained on real and synthetic datasets. Furthermore, the evaluation of submissions is extended to bias assessment across different demography groups. Lastly, an outlook on the current state of the research in training face recognition models using synthetic data is presented, and existing problems as well as potential future directions are also discussed.
\end{abstract}

%%%%%%%%%%%%%%%%%%%%%%%%%%%%%%%%%%%%%%%%%%%%%%%%%%%%%%%%%%%%%%%%%%%%%%%%%%%%%%%%
\section{Introduction}
Recent advancements in state-of-the-art face recognition models are driven in part by the availability of large-scale datasets and deep learning models~\cite{deng2019arcface,kim2022adaface,meng2021magface,george2024edgeface}. Meanwhile, large-scale face recognition datasets, such as VGGFace2~\cite{cao2018vggface2}, MS-Celeb-1M~\cite{guo2016ms}, WebFace260M~\cite{zhu2021webface260m}, etc., were collected by crawling images from the Internet,  raising legal, ethical, and privacy concerns. In light of these concerns and the fact that these datasets are collected without the individuals' consent, 
some dataset owners decided to take their datasets down; for example,  Microsoft retracted MS-Celeb-1M~\cite{Nature_Machine_Intelligence_Datasets}.
To address such concerns, recently, several studies have proposed generating synthetic face recognition datasets and using synthetic face images for training face recognition models~\cite{qiu2021synface,bae2023digiface,colbois2021use,boutros2023idiff}.  
The generated synthetic face recognition datasets should be composed of different synthetic subjects with several samples per subject. 
As a face recognition dataset, each generated synthetic face dataset should include sufficient \textit{intra-class} variations to model challenging scenarios in the task of face recognition including variations caused by pose, aging, expressions, occlusions, environmental conditions, etc. Moreover, the generated dataset needs to have adequate \textit{inter-class} variation so that the trained face recognition models can have generalization for unseen subjects when used in a face recognition system. 
Given these challenges, generating synthetic face datasets with sufficient inter-class and intra-class variations is an active area of research. 

As the main motivation for generating synthetic face recognition datasets, the generation of synthetic face recognition datasets should not rely on large-scale web-crawled datasets which have privacy concerns. In addition, since the synthetic face recognition datasets should eventually replace large-scale real face recognition datasets, face recognition models trained using synthetic datasets should lead to a high recognition accuracy.
However, there is still a gap in the recognition accuracy between training using large-scale real face recognition datasets and existing synthetic face recognition datasets \cite{boutros2023synthetic,shahreza2023synthdistill,melzi2024frcsynongoing}.

The Synthetic Data for Face Recognition (SDFR) competition held in conjunction with the 18th IEEE International Conference on Automatic Face and Gesture Recognition (FG 2024) was organized to accelerate research in synthetic data generation for privacy-friendly face recognition models and to bridge the gap between real and synthetic face datasets. 
Teams were invited to propose clever ways to use synthetic face recognition datasets (either existing or new synthetic face datasets) to train face recognition models. 
The competition was split into two tasks, where the first task involved a predefined face recognition backbone and a limit on the dataset size to focus on the quality of synthesized face datasets, while the second task provided almost complete freedom on the model backbone, the dataset, and the training pipeline. 
The submitted trained face recognition models using synthetic data were evaluated based on a diverse set of seven benchmarking datasets. 
The SDFR competition is the first competition which is specifically focused on generating synthetic face recognition datasets and allows training face recognition models using new methods or existing datasets from the literature. The  competition rules were also designed to allow exploring ideas for generating privacy-friendly datasets, while preventing the application of large-scale web-crawled datasets. 
This paper presents a summary of the  SDFR competition and discusses submissions and findings in this competition.

The remainder of the paper is organized as follows. Section~\ref{section:SDFR_Competition} provides details about the SDFR competition, including the definition of tasks, and evaluation metrics used in the competition. 
In Section~\ref{section:Description_Submissions}, we provide a description of the final submitted face recognition systems proposed in the SDFR competition for each task. Section~\ref{section:Results}  presents the results achieved in the different tasks and compare them with several baselines.  Section~\ref{section:Discussion} extends the analyses for final submission and evaluates their bias on different demography groups.  Section~\ref{section:Discussion} also discusses submissions with not qualified datasets, which are excluded from the leaderboard, and discusses their limitations and findings. 
Furthermore, Section~\ref{section:Discussion} presents a further discussion on training face recognition using synthetic data and highlights potential future research directions in the field. 
Finally, in Section~\ref{section:Conclusion}, we draw the conclusions from the SDFR competition.
 
\section{SDFR Competition}\label{section:SDFR_Competition}
In the SDFR competition, participants were expected to train face recognition models using synthetic data and submit their trained models (in ONNX format)  through a submission platform. The organizers regularly evaluated submitted models on benchmarking datasets, and the results appeared in the leaderboard on the competition website throughout the competition. 
In the following, a brief overview of the definition of competition tasks, rules, and evaluation is presented. Finally, the novelties of the SDFR competition compared to previous competitions on face recognition are discussed.

\subsection{Tasks}
The SDFR competition was divided into two tasks, and each team could use existing synthetic datasets or a newly generated dataset to participate in either or both tasks:
\begin{itemize}
	\item \textbf{Task 1 [Constrained]:} In this task, the generated synthetic dataset could have up to one million synthesized images (for example, 10,000 identities and 100 images per identity). The backbone is also fixed to iResNet-50 \cite{deng2019arcface}.
	\item \textbf{Task 2 [Unconstrained]:} In this task,  participants could use synthetic data with no limit on the number of synthesized images. Participants were also allowed to use any network architecture and train their best model with state-of-the-art techniques, but only using synthetic data.
\end{itemize}
Submitted models in each task were evaluated and ranked in the leaderboard of the corresponding task.

\subsection{Rules}
The competition was organized with several rules which allowed participants to generate new synthetic face recognition datasets and/or use (a subset, the complete set, or an extension of) existing ones to train face recognition models. 
Participants were not allowed to use a real dataset with identity labels for training the face generator model and generate synthetic images. As a result, some of the existing synthetic datasets for the literature, such as  DCFace \cite{kim2023dcface},  SFace \cite{boutros2022sface}, and GANDiffFace~\cite{melzi2023gandiffface}, were not qualified for the competition. Further details about such synthetic datasets are discussed  in Section~\ref{section:Discussion}. 
However, the rules allowed to use real datasets without identity labels (such as FFHQ\footnote{The FFHQ dataset \cite{karras2019style} contains 70,000 images without identity labels with variation in terms of age, ethnicity, etc. Especially, each individual image in this dataset was published on Flickr by their respective authors under licenses that allow free use, redistribution, and adaptation for non-commercial purposes.}~\cite{karras2019style}) to train the face generator model. This could be extended to real datasets with identity labels if the identity labels were not used in training the face generator model\footnote{Given the limitations in available datasets without identity labels, this rule relaxed the restriction in available datasets and allowed participants to use more available datasets. Indeed, this relaxation in the rules was considered to provide the opportunity for participants to investigate limitations in existing datasets without identity labels (such as FFHQ~\cite{karras2019style}). However, outside of the scope of the competition, for a responsible and privacy-friendly synthetic face recognition dataset, large-scale web-crawled datasets (even without labels) should not be used to train generative models.}. 
Participants were also allowed to use a pre-trained face recognition model for controlling and generating synthetic datasets. However, they were not allowed to directly learn embeddings of a pre-trained face recognition model. 
The complete definition of the rules is available on the competition website\footnote{\href{http://www.idiap.ch/challenge/sdfr/}{www.idiap.ch/challenge/sdfr}}.

\subsection{Evaluation}
The submitted models were evaluated on a diverse set of seven benchmarking face recognition datasets. 
The benchmarking datasets include:
\begin{itemize}
	\item \textbf{one high-quality unconstrained  dataset:} Labeled Faces in the Wild (LFW)~\cite{huang2008labeled}, 
	\item \textbf{two cross-pose datasets:}  Celebrities in Frontal-Profile in the Wild (CFP-FP) and Cross-Pose LFW (CPLFW)~\cite{zheng2018cross}, 
	\item \textbf{two cross-age datasets}: AgeDB-30~\cite{moschoglou2017agedb} (30 years age gap) and Cross-age LFW (CALFW)~\cite{zheng2017cross},
	\item \textbf{two challenging mixed-quality datasets:}  IARPA Janus Benchmark–B (IJB-B)~\cite{whitelam2017iarpa} and IARPA Janus Benchmark–C (IJB-C)~\cite{maze2018iarpa}.
\end{itemize}
The submitted models were ranked on each dataset separately, and the final ranking was based on the Borda count.

\subsection{Novelties}
Considering the wide application of face recognition systems and the increasing demand for high-performing face recognition models, several competitions have been organized in conjunction with different conferences~\cite{deng2019lightweight,deng2021masked,boutros2021mfr,kolf2023efar,melzi2024frcsyn}. 
However, most of these competitions were based on large-scale web-crawled face recognition datasets, which raise privacy concerns. 
The FRCSyn challenge held in WACVW 2024 \cite{melzi2024frcsyn,melzi2024frcsynongoing} was the first competition on training face recognition models based on synthetic data; however,  participants were limited to using two existing synthetic datasets from the literature (i.e., DCFace~\cite{kim2023dcface} and GANDiffFace~\cite{melzi2023gandiffface}). 
The SDFR competition is the first competition focused on generating synthetic face recognition datasets, and allows training face recognition models using new methods or existing datasets from the literature (as long as they follow competition rules). The rules of the SDFR competition were also designed to allow exploring ideas for generating privacy-friendly datasets, while preventing the application of large-scale web-crawled datasets.
Furthermore, the SDFR competition is based on a thorough evaluation using seven benchmarking datasets.

\section{Description of Submissions}\label{section:Description_Submissions}
The SDFR competition received significant interest, with 12 international teams registered. 
In total, we received 17 submissions from 5 teams. 
Table~\ref{tab:participants}  provides a summary overview of the participating teams, including the team name, their affiliation, and the tasks they participated. In the following section, we briefly describe the proposed method used by each team.

\textbf{BioLab (Task 1 and Task 2):}
The BioLab team used iResNet-50 \cite{deng2019arcface} for task 1 and an iResNet-100 \cite{deng2019arcface} for task 2.  Both models were trained using the margin-based AdaFace loss\cite{deng2019arcface}, chosen for its state-of-the-art results and robustness during training, especially when employing low-quality images. 
For task 1, they trained their model on the IDiff-Face (Uniform)~\cite{boutros2023idiff} dataset, which contains about 500K images across 10K identities; for task 2, they extended their training dataset in task 1 with the DigiFace-1M~\cite{bae2023digiface} dataset, bringing the total up to 1.7M images spanning 120k classes. 
To improve the performance of the models, they applied data augmentation techniques to the images during training. In particular, they sequentially applied random horizontal flips, color jittering, random crops by blacking out the rest of the image, random JPEG compression, random grayscale, and random down-and-upsampling with different interpolation algorithms (i.e. nearest neighbor, bilinear, bicubic, area, Lanczos). 
When employing images from DigiFace-1M \cite{bae2023digiface}, they applied a more aggressive data augmentation pipeline to bridge the quality gap between synthetic and real images; as suggested in \cite{bae2023digiface}, they extended the above-mentioned pipeline by adding more photometric augmentation steps i.e., Gaussian blur (20\%), random noise (20\%), and motion blur (10\%). 
They built the validation set by generating random matching and non-matching pairs by selecting images from the first classes of IDiff-Face and DigiFace-1M, generating 12K and 84K pairs, respectively, for task 1 and task 2. 
The classes used to generate such pairs are excluded from training. With this validation set, they monitored the verification accuracy during training and applied an early stopping policy with patience of 10 epochs to prevent overfitting. Both models are optimized with SGD  with a batch size of 512, and the initial learning rate of 0.1 divided by a factor of 10 at prefixed epochs to ensure better training stability. 
Code is available: \href{https://github.com/ndido98/sdfr}{https://github.com/ndido98/sdfr}

\begin{table}[tb]
	\centering
	\caption{A summary of the participant teams, their affiliations, and their participating tasks.}
	\label{tab:participants}
	\begin{tabular}{lccc}
		\textbf{Team Name}                                                    & \textbf{Affiliations}$^{\dagger}$ & \textbf{Country} & \textbf{Tasks} \\ \midrule
		BioLab                                                             & 4                   & Italy            & 1,2             \\
		IGD-IDiff-Face                                                             & 5,6                     & Germany              & 1             \\
		APhi                                                    & 7,8                 & Spain           & 1          \\
		BOVIFOCR-UFPR                                                            & 9-11              & Brazil      & 1,2         \\
		BiDA-PRA                                                         & 12,13                 & Spain, Italy           & 1,2       \\
		\bottomrule
		\multicolumn{4}{l}{$^{\dagger}$The numbers reported in the ``Affiliations" column are provided}\\ in the title page.
	\end{tabular}
\end{table}

\textbf{IGD-IDiff-Face (Task 1):}
The submitted solution of the IGD-IDiff-Face team was based on IDiff-Face~\cite{boutros2023idiff}.  The training dataset was obtained by combining two subsets, IDiff-Face CPD25 (Uniform) and IDiff-Face CPD50 (Two-Stage),  of the IDiff-Face dataset. 
These subsets contain 10,049 and 10,050 identities, respectively,  and  50 images per identity. They selected the first exact 10k from each subset. Therefore, the total number of identities is 20K, each contains 50 images. 
They trained the face recognition model using CosFace~\cite{wang2018cosface} with a margin of 0.35 and a scale factor of 64. 
During the training, they augmented the training data using RandAugment~\cite{cubuk2020randaugment} with parameters (operation is 16 and magnitude is 4) described in \cite{boutros2023unsupervised}. 
The model is trained using the SGD optimizer with a batch size of 512. The model is trained using a plateau-based learning scheduler. The initial learning rate is set to 0.1 and divided by 10 when the average validation accuracy does not improve for 10 consecutive epochs where the patience parameter is set to 10 and the factor to 0.4. The total number of epochs is set to 200 with an early stopping parameter of 10 epochs, i.e., the training will stop if the average validation accuracy does not improve for 10 epochs. Code is available: \href{https://github.com/fdbtrs/IDiff-Face}{https://github.com/fdbtrs/IDiff-Face}

\textbf{APhi (Task 1):}
The APhi team used iResNet-50 as the backbone of their model and trained it with MagFace~\cite{meng2021magface} loss function.
As their training dataset, they used the IDiff-Face  (Uniform) \cite{boutros2023idiff}, which consists of about 550K images and 10K identities.
They used data augmentation techniques, including random scale and then resize, Gaussian blur, Hue saturation, and horizontal flip.
They trained their model for 25  epochs using the  SGD  optimizer with a batch size of 512 and a learning rate initialized from  0.1  and divided by  10  at epochs  10,  18, and  22.  The weight decay was set to  $5\times10^{-4}$  and the momentum is 0.9. In addition, 200 users have been taken from the training set and used for validation (a total of  10K  images). The validation was performed by evaluating the equal error rate (EER) over all the possible pairs in the validation set. They validated the trained model every 1,000 iterations, and stored the model with the lowest EER.

\textbf{BOVIFOCR-UFPR (Task 1 and Task 2):}
The BOVIFOCR-UFPR team used iResNet-50 \cite{deng2019arcface}  for both task 1 and task 2 and trained their models using  ArcFace loss~\cite{deng2019arcface}, chosen for its performance for deep face recognition \cite{deng2021masked}. 
For task 1, they used the uniform version of the IDiff-Face \cite{boutros2023idiff} as the training dataset containing $10,049$ identities, with $55$ images per identity  ($502,450$ images in total). 
The  training images were augmented using Random Flip with a probability of
$50\%$. 
They also applied random erasing \cite{zhong2020random} and randaugment \cite{cubuk2020randaugment} as additional augmentation. 
They used the SGD optimizer and set the momentum to $0.9$ and weight decay to $5\times10^{-4}$. The learning rate was set to $0.02$ and reduced at each iteration following the equation:

\begin{equation}\label{eq:BOVIFOCR-lr}
	\left( \frac{1.0 - \frac{l}{t}}{1.0 - \frac{l - 1}{t}} \right),
\end{equation}

where $l$ and $t$ represent the current iteration and the total number of iterations, respectively.

The model was trained for $20$ epochs within a batch size of $128$, running for approximately $78$K iterations  using the Insightface library\footnote{\href{https://github.com/deepinsight/insightface}{https://github.com/deepinsight/insightface}}. Code for task 1 is available: \href{https://github.com/PedroBVidal/insightface}{https://github.com/PedroBVidal/insightface}

\begin{figure}[tb]
		\includegraphics[width=1\linewidth, trim={0cm 0cm 0cm 0.5cm},clip]{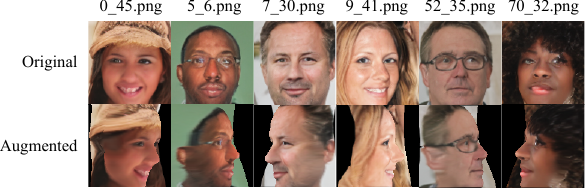}
	\caption{Sample original and augmented face images from the IDiff-Face (Uniform) dataset obtained by synthetically rotating face pose in the yaw axis used by the BOVIFOCR-UFPR team in task 2.}
	\label{fig:BOVIFOCR-UFPR:face_pose_aug}
\end{figure}

For task 2, they also used IDiff-Face (Uniform) \cite{boutros2023idiff}, and applied random flip and also random face pose augmentation \cite{cheng2020towards} based on 3D face reconstruction \cite{zhu2017face} to manipulate face pose in yaw axis. About $40$K faces were randomly profiled in left ($-60^{\circ}$) or right ($+60^{\circ}$) direction, corresponding to $\mathtt{\sim}8\%$ from the whole dataset.  Some sample augmented face images are shown in Fig. \ref{fig:BOVIFOCR-UFPR:face_pose_aug}. 
The used  SGD optimizer with learning rate $0.1$, momentum $0.9$, batch size $256$,  and  weight decay of $5\times10^{-4}$.
The model was trained during $20$ epochs, corresponding to $84$k iterations. 
Code for task 2 is also available: \href{https://github.com/BOVIFOCR/insightface_SDFR_at_FG2024}{https://github.com/BOVIFOCR/insightface\_SDFR\_at\_FG2024}

\begin{table*}
	\centering
	\renewcommand{\arraystretch}{1.275}
	\caption{Leaderboard of Task 1 and Task 2 compared to several baselines.}
	\label{tab:leaderboard}
	\resizebox{\linewidth}{!}{%
		\begin{tabular}{|c|c|c|c|c|c|c|c|c|c|c|} 
			\hline
			\vspace{-3pt}\multirow{2}{*}{\textbf{Task / Baseline}}               & \multirow{2}{*}{\textbf{Team / Method }}  & \multirow{2}{*}{\textbf{No. Images}} &  \multirow{2}{*}{\textbf{LFW}} &  \multirow{2}{*}{\textbf{CALFW}} & \multirow{2}{*}{\textbf{CPLFW}}	& \multirow{2}{*}{\textbf{AgeDB30}} & \multirow{2}{*}{\textbf{CFP-FP}} & \textbf{IJB-B} & \textbf{IJB-C} & \multirow{2}{*}{\textbf{Rank}}  \\ 
			& & &  & & & & & \scalebox{0.9}{\textbf{(TAR@$\mathbf{10^{-4}}$)}} & \scalebox{0.9}{\textbf{(TAR@$\mathbf{10^{-4}}$)}} & \\
			\hline 
			%baselines (real)
			\hline
			\multirow{2}{*}{Baselines}    & MS-Celeb \cite{guo2016ms} & 5.8M & 99.82 & 95.92 & 92.52 & 97.62 & 96.01 &   94.88 & 96.23 &  -     \\ \cline{2-11}
			 \multirow{2}{*}{(real)}       & WebFace-4M  \cite{zhu2021webface260m}  & 4M &  99.78 & 96.02 & 93.90 & 97.52 & 97.19 &   95.52 & 97.02  &  -      \\ \cline{2-11}
			   & Casia-WebFace \cite{yi2014learning}   & 490K &     99.08 & 92.88 & 89.23 & 93.72 & 94.67 & 45.91  &  52.43  &  -        \\ 
			\hline 
			%baselines (synthetic)
			\hline
			\multirow{2}{*}{Baselines}             & SynFace~\cite{qiu2021synface}            & 1M  &  85.55 & 69.85 & 59.42 & 58.38 & 62.11 & 14.62 & 12.85 & -  \\ \cline{2-11}
			\multirow{2}{*}{(synthetic)}       & DigiFace~\cite{bae2023digiface}            & 1.2M  & 90.63 & 74.02 & 71.38 & 65.03 & 78.11 & 38.89 & 45.09 & -  \\  \cline{2-11}
			      & IDNet~\cite{kolf2023identity}            & 1.05M  &  84.90 & 70.28 & 68.32 & 63.88 & 70.34 &  27.57 & 32.28 & -\\ 
			\hline 
			%Task1
			\hline
			\multirow{5}{*}{Task 1}             & IGD-IDiff-Face            & 1M     &  98.07 & 90.60 & 81.28 & 87.60 & 84.76  & 64.36 & 68.04 & 1  \\ \cline{2-11}
			\multirow{5}{*}{(leaderboard)}    & APhi   & 500K & 97.45 & 89.95 & 78.03 & 84.75 & 80.04 & 58.18 & 60.85 & 2\\ \cline{2-11}
			& BOVIFOCR-UFPR            & 500K   & 97.53 & 89.38 & 80.07 & 83.90  & 84.37 & 12.70 & 13.71 & 3  \\ \cline{2-11}
			& BioLab            & 500K    & 96.97 &  89.12 &  76.80 & 83.77 &  77.34 &  60.21 & 63.56 &  4  \\ \cline{2-11}
			& BiDA-PRA     & 1M   & 96.88 & 	87.95 &  78.13 &  83.85 & 78.90 & 58.08 & 58.33 &  5 \\ 
			\hline 
			%Task 2
			\hline
			\multirow{2}{*}{Task 2}            
			& BioLab            & 1.7M   & 98.33 & 90.87 & 84.45 & 87.85 & 88.11 & 76.94 & 81.25 & 1  \\ \cline{2-11}
			\multirow{2}{*}{(leaderboard)}
			& BiDA-PRA    & 1M   & 96.88 & 	87.95 &  78.13 &  83.85 & 78.90 & 58.08 & 58.33 & 2 \\  	\cline{2-11}
			& BOVIFOCR-UFPR          & 500K  & 96.38 & 88.38 & 75.98 & 81.55 & 76.97 & 40.97 & 45.93 & 3 \\  \hline 
			\multicolumn{11}{l}{\vspace{-2pt}The values reported in the table for each benchmarking dataset are in percentage. The values reported on the LFW, CALFW, CPLFW, AgeDB30, and  CFP-FP   }\\
			\multicolumn{11}{l}{datasets are accuracy. The values reported  on the IJB-B and  IJB-C datasets are True Accept Rate (TAR) at the False Accept Rate (FAR) of $10^{-4}$.}\\
			\multicolumn{11}{l}{\vspace{-2pt}All baseline models are trained with iResNet-50 and AdaFace  \cite{kim2022adaface} loss function.
				Baselines with real data are pretrained models available on the AdaFace  }\\
			\multicolumn{11}{l}{GitHub repository, and baseline models with synthetic data are trained by organizers using the same hyperparamters. 
				}
			
		\end{tabular}
	}
\end{table*}

\textbf{BiDA-PRA (Task 1 and Task 2):}
The BiDA-PRA  team submitted the same model for both tasks. They used iResNet-50 \cite{deng2019arcface} and trained using the AdaFace \cite{kim2022adaface} loss function. 
For preparing their training dataset, they used the information about demographic distribution across the competition evaluation datasets, focusing on ethnicity, gender, and age, available from the datasets meta-data or from the analyses reported in previous studies, such as \cite{kong2021rsfad}.  
For the training dataset, they used pre-generated IDiff-Face synthetic dataset (both Uniform and Two-stage subsets, containing about 1M images in total), and created a ``cleaned" version of the dataset, discarding each identity image whose labels were different from the average ethnicity and gender predictions.  
These attributes were automatically extracted for each face image in the IDiff-Face dataset using the FairFace~\cite{karkkainen2021fairface} model. 
Finally, after the cleaning step, they considered 600K total face images of the IDiff-Face dataset. In addition, to generate more training samples, they trained\footnote{The BiDA-PRA  team considered that pre-trained  StyleGAN3 \cite{karras2021alias}   with the FFHQ  \cite{karras2019style} dataset exhibit superior quality compared to the images in the evaluation datasets. Therefore, they trained StyleGAN3 \cite{karras2021alias} on the CASIA-WebFace \cite{yi2014learning} dataset, which consists of images with more similar quality compared to evaluation datasets.} StyleGAN3 on CASIA-WebFace \cite{yi2014learning} (without using its identity labels) and used it to generate 20K images of unique identities. They applied the method in \cite{melzi2023gandiffface} to generate pose variations in the latent space of those images, resulting in 20 images per identity, and in total 400K face images generated by  StyleGAN3. However, they did not perform an analysis of the demographic distribution for the images generated by  StyleGAN3. 
During training, they augmented data using random histogram equalization (10\%), random horizontal flip (50\%), random grayscale conversion (20\%), and Gaussian blur (80\% with kernel 1, and 20\% with kernel 5). 
In addition, they applied a custom augmentation technique involved cutting off half of the face, either the left or the right side, from some of the frontal or near-frontal images\footnote{This custom augmentation technique was intended to train the network to extract useful information from half of the face, as in the case of profile images.}.  
In particular, for each selected image, they checked if the face could be considered frontal or not through a landmark-based estimation.  If the image was considered frontal, they randomly selected the left or the right side to be removed.  They detected the landmarks of the nose and the eyes using MTCNN \cite{zhang2016joint}, and then they computed the difference between the angles on the eyes from the triangle composed of these three landmarks.   Specifically,  they selected a maximum angle difference of  15 degrees to consider the face as frontal or near-frontal, while setting the image selection probability to 5\%. Code is available: \href{https://github.com/BiDAlab/SDFR-FRModels}{https://github.com/BiDAlab/SDFR-FRModels}

\section{Leaderboards}\label{section:Results}
As mentioned in Section~\ref{section:SDFR_Competition}, submitted models are evaluated on a diverse set of benchmarking datasets. Table~\ref{tab:leaderboard}  reports the leaderboard of each task and also compares the results of submissions with some baselines from the literature. The values in the table for each benchmarking dataset are in percentage. The values reported on the LFW, CALFW, CPLFW, AgeDB30, and  CFP-FP datasets are accuracy and the values reported on the IJB-B and  IJB-C datasets are True Accept Rate (TAR) at the False Accept Rate (FAR) of $10^{-4}$.
The baselines were trained by the corresponding dataset mentioned in the table using iResNet-50 \cite{deng2019arcface} network with  AdaFace~\cite{kim2022adaface} loss function. 
As can be seen in Table~\ref{tab:leaderboard}, there is a gap in the recognition performance of baselines with real datasets and baselines with synthetic datasets. This difference is more significant for challenging datasets such as IJB-B, and IJB-C. However, the submitted models could improve the performance compared to baselines with synthetic datasets. In particular, the submitted model by the BioLab team for task 2 achieved considerable performance on all datasets, even on challenging datasets such as IJB-B and IJB-C. This submission used two different datasets, DigiFace~\cite{bae2023digiface} (a synthetic dataset based on computer graphic pipeline) and IDiff-Face~\cite{boutros2023idiff} (a synthetic dataset based on diffusion models), showing potentials in improving performance using synthetic datasets with different types. 
While task 1 received significant interest from the participating teams, task 2 received fewer submissions, indicating the difficulty of task 2 which allowed unconstrained training with synthetic data. 

In task 1,  the IGD-IDiff-Face team used two subsets of IDiff-Face (Uniform and Two-stage), resulting in larger training data, compared to the APhi, BOVIFOCR-UFPR, and BioLab teams, who used only the Uniform subset of IDiff-Face with about 500K images. Meanwhile, the submission of BiDA-PRA  also benefits from both subsets of IDiff-Face and in addition used StyleGAN generated images, using the complete limit of 1M images in task 1. While using 1M training samples helped the performance of their submission by the IGD-IDiff-Face team, it could not help the performance of the BiDA-PRA  team. This might be caused by not enough variation in their StyleGAN-generated images, which only had pose variation.

All the teams also used different data augmentation techniques, to further improve their trained models. The data augmentation transformations used by participants contain various techniques and include traditional methods (such as random flip, Gaussian blur, etc), state-of-the-art techniques (such as random erasing \cite{zhong2020random}, randaugment \cite{cubuk2020randaugment},  and pose augmentation \cite{cheng2020towards}), as well as custom methods (such as cutting off half of the face). This highlights the importance of data augmentation in training with synthetic data, which lack intra-class variations. 

Last but not least, all final submissions have used the IDiff-Face dataset as part of their training, with different sampling strategies and training approaches. In fact, the IDiff-Face dataset is the best-performing synthetic dataset in the literature which is aligned with the competition rules. We should note that participating teams also submitted other models, which had competitive performance with the final leaderboard but had a conflict with the competition rules. Such submissions with not qualified datasets are discussed in detail in Section~\ref{section:Discussion}.
To facilitate reproducibility of results in the SDFR competition, all final submission which appeared on the final leaderboard are publicly available on the competition website: \href{http://www.idiap.ch/challenge/sdfr/}{www.idiap.ch/challenge/sdfr}

\section{Discussion}\label{section:Discussion}
In addition to our evaluation of submitted models in Table~\ref{tab:leaderboard}, we evaluated each of the models in the leaderboard on the Racial Faces in-the-Wild (RFW)  \cite{wang2019racial} dataset, to investigate their performance across different demography groups. The RFW dataset contains four testing subsets corresponding to Caucasian, Asian, Indian, and African groups, and each is composed of about 3,000 subjects with 6,000 image pairs for the face verification task. 
Table~\ref{tab:RFW} reports the performance of the final submissions in the leaderboard  as well as baseline models trained on real and synthetic datasets on the different subsets of the RFW dataset. As the results in this table show, all final submissions from the leaderboard achieved their best performance in the Caucasian group and worst performance in the African group. In each task, the best submission in the leaderboard in Table~\ref{tab:leaderboard} also achieves the highest average and lowest standard deviation across different  demography groups on the RFW dataset.  The results in Table~\ref{tab:RFW} also show  that there is similarly bias in the baselines trained on real and synthetic datasets. 
Hence, the results in this table motivates future research on generating responsible synthetic face recognition datasets to alleviate bias in recognition accuracy across different  demography groups.

\begin{table}
	\centering
	\renewcommand{\arraystretch}{1.3}
	\setlength{\tabcolsep}{2.45pt}
	\caption{Evaluation results of the final submissions on the RFW dataset.}
	\label{tab:RFW}
	\resizebox{\linewidth}{!}{%
		\begin{tabular}{|c|c|c|c|c|c|c|c|c|c|}
			\hline 
			 \scalebox{0.9}{\textbf{Task\scalebox{0.25}{ }/\scalebox{0.25}{ }Baseline} }              & \scalebox{0.975}{\textbf{Team\scalebox{0.5}{ }/\scalebox{0.5}{ }Method}}  & \scalebox{0.9}{\textbf{Caucasian}} &  \textbf{Asian} & \scalebox{0.975}{\textbf{Indian}} & \scalebox{0.95}{\textbf{African}} & \textbf{Avg.} & \scalebox{0.975}{\textbf{Std.}}  \\ 
			\hline
			\hline 
\multirow{2}{*}{Baselines}            
& MS-Celeb \cite{guo2016ms}             &      99.33 & 97.73 & 98.23 & 98.32 &  98.40 &  0.68   \\ \cline{2-8}
\multirow{2}{*}{(real)}            
& WebFace-4M  \cite{zhu2021webface260m}   &   98.98 & 97.10 & 97.72 & 97.63 &  97.86 &  0.80   \\  	\cline{2-8}
& \scalebox{0.95}{Casia-WebFace \cite{yi2014learning}}    &    93.05 & 85.07 & 88.20 & 85.57 &  87.97 & 3.66  \\  
\hline 
\hline 
\multirow{2}{*}{Baselines}            
& SynFace \cite{qiu2021synface}    &      65.60 & 64.48 & 61.48 & 57.27 &  62.21 & 3.73   \\ \cline{2-8}
\multirow{2}{*}{(synthetic)}            
& DigiFace  \cite{bae2023digiface}   &    71.93 & 68.30 & 69.02 & 64.80 &  68.52 &  2.94   \\  	\cline{2-8}
& IDNet \cite{kolf2023identity}     &      70.03 & 64.22 & 65.77 & 59.30 &  64.83 &  4.44  \\  
\hline 
\hline 
\multirow{5}{*}{Task 1}             
& IGD-IDiff-Face    &  85.98 & 80.98 & 82.45 & 76.85 & 81.57 &  3.78  \\ \cline{2-8}
& APhi   & 84.55 & 79.45 & 80.70 & 73.85 & 79.64 &  4.43 \\ \cline{2-8}
& \scalebox{0.95}{BOVIFOCR-UFPR}   & 83.65 & 78.92 & 80.18 & 74.72 & 79.37 &  3.69 \\ \cline{2-8}
& BioLab                &   81.70 & 77.22 & 78.97 & 71.20 & 77.27 & 4.45 \\ \cline{2-8}
& BiDA-PRA    &   82.12 & 75.87 & 77.88 & 70.00 & 76.47 & 5.04 \\ 
\hline 
\hline 
%Task 2
\multirow{3}{*}{Task 2}            
& BioLab            &  86.32 & 81.47 & 82.22 & 77.30 & 81.82 & 3.70 \\ \cline{2-8}
& BiDA-PRA    &   82.12 & 75.87 & 77.88 & 70.00 & 76.47 & 3.48 \\  	\cline{2-8}
& \scalebox{0.95}{BOVIFOCR-UFPR}    & 80.37 & 76.47 & 77.65 & 71.08 & 76.39 & 3.91\\  \hline 
\multicolumn{8}{l}{The values reported in the table are accuracy and in percentage.  }\\
		\end{tabular}
	}
\end{table}

Some of the synthetic datasets in the literature, such as DCFace \cite{kim2023dcface},  SFace \cite{boutros2022sface}, and GANDiffFace~\cite{melzi2023gandiffface}, have a conflict with the competition rules, and therefore participants were not allowed to use them for their final submission. 
DCFace \cite{kim2023dcface} used CASIA-WebFace \cite{yi2014learning} to train the dual condition generator, where training the generator requires two samples of the same subject, and thus it requires an identity-labeled dataset.  SFace \cite{boutros2022sface} also used CASIA-WebFace with identity labels to train a conditional version of  StyleGAN2-ADA \cite{karras2020training} (conditioned on identity). Therefore, DCFace and SFace rely on CASIA-WebFace, as a large-scale web-crawled dataset. 
GANDiffFace \cite{melzi2023gandiffface} used StyleGAN to generate different synthetic images and then used DreamBooth \cite{ruiz2023dreambooth} to generate different samples of images that were generated by StyleGAN in the first place. However, DreamBooth \cite{ruiz2023dreambooth} is based on Stable Diffusion \cite{rombach2022high}, which is trained on the LAION \cite{schuhmann2022laion} dataset. The LAION dataset is a large-scale web-crawled dataset, which includes the identity label of famous people in the caption of images. As a result, Stable Diffusion (a text-to-image model) can generate images of famous people given the name of the person in the input text \cite{yang2024implicit}, raising privacy and ethical issues. Therefore, since the generator in GANDiffFace is trained and benefits from identity labels in the LAION dataset, the GANDiffFace dataset was not allowed in the competition.

\begin{figure}[tb]
	\captionsetup[subfigure]{aboveskip=0.6pt}	\captionsetup[subfigure]{belowskip=-0.8pt}
	\centering
	\begin{subfigure}[b]{0.19\linewidth}
		\centering
		\includegraphics[page=1,width=.95\linewidth, trim={0.1cm 0cm 0cm 0cm},clip]{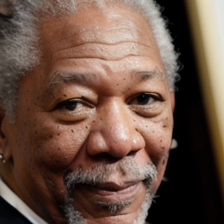}
	\end{subfigure}\hfil
	\begin{subfigure}[b]{0.19\linewidth}
		\centering
		\includegraphics[page=1,width=0.95\linewidth, trim={0.1cm 0cm 0cm 0cm},clip]{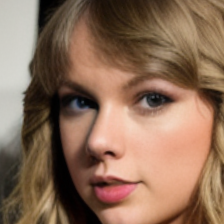}
	\end{subfigure}\hfil
	\begin{subfigure}[b]{0.19\linewidth}
		\centering
		\includegraphics[page=1,width=0.95\linewidth, trim={0.1cm 0cm 0cm 0cm},clip]{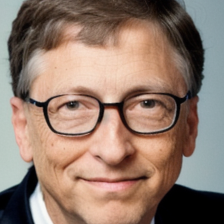}
	\end{subfigure}\hfil
	\begin{subfigure}[b]{0.19\linewidth}
		\centering
		\includegraphics[page=1,width=0.95\linewidth, trim={0.1cm 0cm 0cm 0cm},clip]{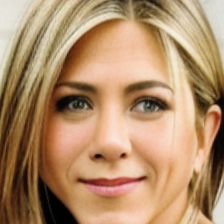}
	\end{subfigure}\hfil 
	\begin{subfigure}[b]{0.19\linewidth}
	\centering
	\includegraphics[page=1,width=0.95\linewidth, trim={0.1cm 0cm 0cm 0cm},clip]{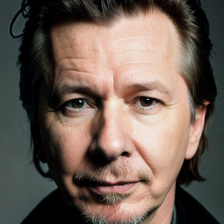}
	\end{subfigure}\hfil \\ \vspace{3pt}
	\begin{subfigure}[b]{0.19\linewidth}
		\centering
		\includegraphics[page=1,width=.95\linewidth, trim={0.1cm 0cm 0cm 0cm},clip]{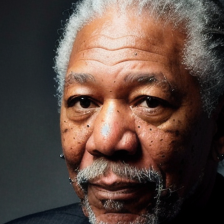}
	\end{subfigure}\hfil
	\begin{subfigure}[b]{0.19\linewidth}
		\centering
		\includegraphics[page=1,width=.95\linewidth, trim={0.1cm 0cm 0cm 0cm},clip]{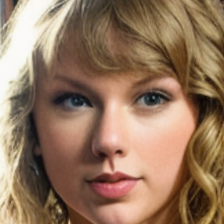}
	\end{subfigure}\hfil
	\begin{subfigure}[b]{0.19\linewidth}
		\centering
		\includegraphics[page=1,width=.95\linewidth, trim={0.1cm 0cm 0cm 0cm},clip]{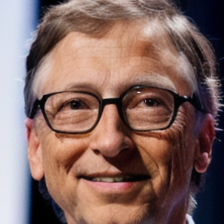}
	\end{subfigure}\hfil
	\begin{subfigure}[b]{0.19\linewidth}
		\centering
		\includegraphics[page=1,width=0.95\linewidth, trim={0.1cm 0cm 0cm 0cm},clip]{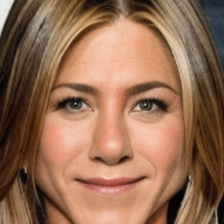}
	\end{subfigure}\hfil
	\begin{subfigure}[b]{0.19\linewidth}
	\centering
	\includegraphics[page=1,width=0.95\linewidth, trim={0.1cm 0cm 0cm 0cm},clip]{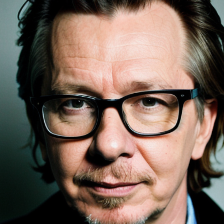}
	\end{subfigure}\hfil\\ \vspace{3pt}
	\begin{subfigure}[b]{0.19\linewidth}
		\centering
		\includegraphics[page=1,width=.95\linewidth, trim={0.1cm 0cm 0cm 0cm},clip]{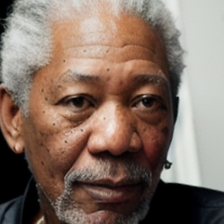}
		\caption{}
	\end{subfigure}\hfil
	\begin{subfigure}[b]{0.19\linewidth}
		\centering
		\includegraphics[page=1,width=.95\linewidth, trim={0.1cm 0cm 0cm 0cm},clip]{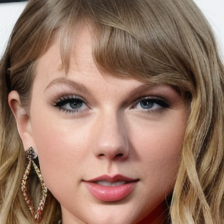}
		\caption{}
	\end{subfigure}\hfil
	\begin{subfigure}[b]{0.19\linewidth}
		\centering
		\includegraphics[page=1,width=.95\linewidth, trim={0.1cm 0cm 0cm 0cm},clip]{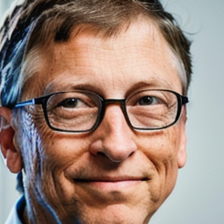}
		\caption{}
	\end{subfigure}\hfil
	\begin{subfigure}[b]{0.19\linewidth}
		\centering
		\includegraphics[page=1,width=0.95\linewidth, trim={0.1cm 0cm 0cm 0cm},clip]{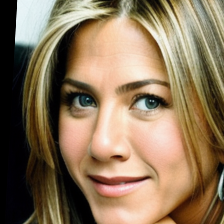}
		\caption{}
	\end{subfigure}\hfil
	\begin{subfigure}[b]{0.19\linewidth}
	\centering
	\includegraphics[page=1,width=0.95\linewidth, trim={0.1cm 0cm 0cm 0cm},clip]{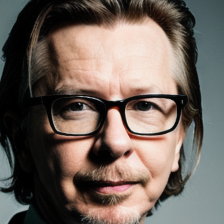}
	\caption{}
	\end{subfigure}\hfil\\ \vspace{0 pt}
	\caption{Sample face images from the new custom generated dataset by the BioLab team generated for different famous people: 
		(a) Morgan Freeman, (b) Taylor Swift, (c) Bill Gates, (d) Jennifer Aniston, (e) Gary Oldman.
	}
	\label{fig:sample_BioLab}
\end{figure}

In addition to the submissions reported in the leaderboard, some participating teams trained models with datasets which had a conflict with the competition rules. 
The  BOVIFOCR-UFPR team also used DCFace \cite{kim2023dcface} with 1.2M images, and dropped 5,454 identities with 55 images per id, reducing the total number of images to 999,975 to fulfill the maximum allowed number of images in task 1. Then, they trained iResNet-50 with the same data augmentation and loss function as their final submission for task 1 described in Section~\ref{section:Description_Submissions}.
The APhi team submitted two models for task 1 trained on DCFace \cite{kim2023dcface}. In the first model, they selected arbitrary 1M images with 59,968 identities and trained the iResNet-50 network with AdaFace \cite{kim2022adaface} loss function. In their second model, they selected 1M based on the face quality. To this end,  they extracted the  MagFace \cite{meng2021magface} score of each image,  and then they selected the best quality images, resulting in 1M images and 59,894 identities. Then, they trained the iResNet-50 network with ArcFace~\cite{deng2019arcface} loss function.
The BioLab team submitted a model trained with a newly generated dataset based on Stable Diffusion as well as with DCFace.
Their newly generated dataset based on Stable Diffusion, was built by querying Wikidata\footnote{\href{https://www.wikidata.org/}{www.wikidata.org}} for all people born between 1900 and 2004 that have an associated picture. To remove subjects that may be difficult to generate with Stable Diffusion, they filtered the list by keeping only the ones with at least 30 translations of their Wikipedia page. After gathering the list, they generated 64 images per subject by employing the Realistic Vision 5.1 model, fine-tuned from Stable Diffusion 1.5. Then, they filtered the generated images by computing ArcFace \cite{deng2019arcface} features for each image, computing the distance between each pair of images of each subject, and finally keeping the largest subset so that the cosine distance between all feature vectors is less than a given threshold. Moreover, if after this filtering step, there were not enough images for a given subject, that subject was removed entirely from the dataset.
This process led to a dataset with around 825K images across 18K identities.  Fig.~\ref{fig:sample_BioLab} illustrates sample images from their generated dataset.
For task 1, they used their newly generated dataset\footnote{As this approach relies on Stable Diffusion (trained on the LAION~\cite{schuhmann2022laion} dataset) for image generation, this submission had a conflict with competition rules, which prevent the use of large-scale datasets with identity labels to train face generator network.} and trained iResNet-50 with the same data augmentation and loss function as their final submission described in Section~\ref{section:Description_Submissions}. For task 2, they also used the same training approach to train the iResNet-100 network, but using the DCFace dataset with 1.2M images. 
Table~\ref{tab:not-qualified-submissions} reports the performance of these submissions on benchmarking datasets. As the results in this table show, these submissions achieve competitive performance with the final submissions on the leaderboards in Table~\ref{tab:leaderboard}.

\begin{table}
	\centering
	\renewcommand{\arraystretch}{1.25}
	\setlength{\tabcolsep}{2.35pt}
	\caption{Submissions with not qualified datasets.}
	\label{tab:not-qualified-submissions}
	\resizebox{\linewidth}{!}{%
		\begin{tabular}{|l|c|c|c|c|c|c|c|c|c|c|} 
			\hline
			\vspace{-3pt}\multirow{2}{*}{\textbf{Team (method) }}  &  \scalebox{0.85}{\multirow{2}{*}{\textbf{LFW}}} &   \scalebox{0.825}{\multirow{2}{*}{\textbf{CALFW}}} & \scalebox{0.825}{ \multirow{2}{*}{\textbf{CPLFW}}}	&  \scalebox{0.825}{\multirow{2}{*}{\textbf{AgeDB30}}} &  \scalebox{0.825}{\multirow{2}{*}{\textbf{CFP-FP}}} &  \scalebox{0.85}{\textbf{IJB-B}} &  \scalebox{0.85}{\textbf{IJB-C}} \\ 
			&   & & & & & \scalebox{0.685}{\textbf{(TAR@$\mathbf{10^{-4}}$)}} & \scalebox{0.685}{\textbf{(TAR@$\mathbf{10^{-4}}$)}}  \\
			\hline 
			\scalebox{0.925}{BOVIFOCR-UFPR}   & 98.98 & 92.42 & 86.38 & 91.38 & 90.89 & 79.62 & 84.11  \\  \hline 		
			APhi \scalebox{0.95}{(model 1) }  & 98.83 & 92.25 & 82.93 & 90.95 & 87.36 & 78.41 &82.38  \\ \hline 
			APhi \scalebox{0.95}{(model 2) }   & 98.38 & 92.22 & 84.27 & 90.88 & 87.57 &  79.48 & 83.61 \\  \hline 	 
			\vspace{-4pt} \multirow{1}{*}{BioLab }   &  \multirow{2}{*}{ 93.90} &  \multirow{2}{*}{82.30} &  \multirow{2}{*}{71.23} &  \multirow{2}{*}{74.42} &  \multirow{2}{*}{72.04} &  \multirow{2}{*}{49.87} & \multirow{2}{*}{52.76} \\ 
			\scalebox{0.95}{(custom dataset)} &   & & & & & & \\ \hline 
			BioLab \scalebox{0.95}{(DCFace)} & 98.52 & 92.48 & 83.35 & 91.12 & 87.63 & 30.17 & 24.38  \\  \hline 
		\end{tabular}
	}
\end{table}

In drawing our discussion to a close, the submissions in this competition used creative approaches for training face recognition using synthetic data and could achieve improvements compared to baselines based on synthetic datasets. However, as the results in Table \ref{tab:leaderboard} show, we still observe a significant gap between models trained with synthetic data and models trained with large-scale web-crawled datasets, demanding more research on generation and training face recognition with synthetic data. In fact, one aspect of this challenge is to increase inter-class and intra-class variation in generating synthetic face datasets. Compared to large-scale datasets, such as MS-Celeb which has 5.8M images, the largest synthetic face recognition dataset in the literature has 1.2M images (DigiFace). Therefore, another potential aspect of the topic for future work can be to explore if we can scale synthetic face datasets to generate more images and investigate if it can contribute to the recognition performance of face recognition models. In fact, scaling synthetic face recognition datasets may not be straightforward, as we may achieve the generation capacity of face generator networks~\cite{boddeti2023biometric}, which also requires further studies.
We should also note that in this competition and in most methods for generating synthetic face recognition datasets, still a pre-trained face recognition model is used in the dataset generation process. Therefore, if we agree on using a pre-trained face recognition model in data generation, it has been shown that directly learning embeddings of a pre-trained face recognition model using synthetic data, without any identity class, achieves superior performance than training using existing synthetic face recognition datasets  \cite{shahreza2023synthdistill}. 
However,
in a \textit{fully} privacy-friendly and responsible synthetic face recognition dataset, no pre-trained face recognition model based on large-scale web-crawled face datasets should be used, otherwise, the generated dataset still relies on large-scale web-crawled datasets. This is in fact an important future research direction that requires more attention from the research community.

\section{Conclusion}\label{section:Conclusion}
Existing large-scale face recognition datasets are collected without individuals' consent and by crawling the Internet, raising legal, ethical, and privacy concerns. While generating a synthetic face recognition dataset can be a potential solution, it is challenging to generate a synthetic dataset with sufficient inter-class and intra-class variations. As a result, there is still a gap between training face recognition models with real and synthetic datasets. The submitted models in the SDFR competition deployed different techniques using newly generated and existing datasets to improve the performance of face recognition models trained with synthetic datasets. 
In this paper, different submissions from participating teams in the SDFR competition were presented and their trained models were evaluated on a diverse set of benchmarking datasets. We also outlined the current state of research on this topic and discussed open problems, such as scaling synthetic datasets as well as increasing the variations in generated images, which require more research in the future.

%%%%%%%%%%%%%%%%%%%%%%%%%%%%%%%%%%%%%%%%%%%%%%%%%%%%%%%%%%%%%%%%%%%%%%%%%%%%%%%%
\section*{Acknowledgments}
The organization of this competition was supported by the H2020 TReSPAsS-ETN Marie Sk\l{}odowska-Curie early training network (grant agreement 860813) as well as the Hasler foundation through the ``Responsible Face Recognition" (SAFER)  project.

The work of BioLab team received funding from the European Union’s Horizon 2020 research and innovation program under Grant Agreement No. 883356 (Disclaimer: the text reflects only the author’s views, and the Commission is not liable for any use that may be made of the information contained therein). The BioLab team would like to thank Andrea Pilzer, NVIDIA AI Technology Center, EMEA, for his support. The BioLab team also acknowledge the CINECA award under the ISCRA initiative, for the availability of high-performance computing resources and support.

The submitted solution by the IGD-IDiff-Face team has been funded by the German Federal Ministry of Education and Research and the Hessian Ministry of Higher Education, Research, Science and the Arts within their joint support of the National Research Center for Applied Cybersecurity ATHENE.

The work of BiDA-PRA  team was supported by INTER-ACTION (PID2021-126521OB-I00 MICINN/FEDER), Cátedra ENIA UAM-VERIDAS en IA Responsable (NextGenerationEU PRTR TSI-100927-2023-2), and R\&D Agreement DGGC/UAM/FUAM for Biometrics and Cybersecurity.
The work of BiDA-PRA  team was also supported by the European Union – Next Generation EU through the Italian Ministry of University and Research (MUR) within the PRIN PNRR 2022 – BullyBuster 2 - the ongoing fight against bullying and cyberbullying with the help of artificial intelligence for the human wellbeing (CUP: F53D23009240001).

%%%%%%%%%%%%%%%%%%%%%%%%%%%%%%%%%%%%%%%%%%%%%%%%%%%%%%%%%%%%%%%%%%%%%%%%%%%%%%%%

{\small
\bibliographystyle{ieee}
\bibliography{egbib}
}

\end{document}